\documentclass{article}
\usepackage{amsmath,graphicx,mlspconf}
\usepackage{xcolor}

\copyrightnotice{979-8-3503-2411-2/25/\$31.00 {\copyright}2025 European Union}

\toappear{2025 IEEE International Workshop on Machine Learning for Signal Processing, Aug.\ 31-- Sep.\ 3, 2025, Istanbul, Turkey}
\usepackage{multirow}
\usepackage{booktabs,tabularx}
\usepackage{tipa}
\usepackage[
    backend=biber,
    style = ieee,
]{biblatex}

\usepackage{url}

\usepackage{breakurl}
\usepackage[breaklinks]{hyperref}
\title{Non-native Children's Automatic Speech Assessment Challenge (NOCASA)}
\twoauthors
 {Yaroslav Getman, Tam\'as Gr\'osz, Mikko Kurimo }
 {Department of Information\\ and Communications Engineering,\\ Aalto University, Finland}
  {Giampiero Salvi}
   {Department of Electronic Systems, \\
   NTNU, Norway}
\begin{document}
%
\maketitle
\begin{abstract}

This paper presents the "Non-native Children's Automatic Speech Assessment" (NOCASA)  -- a data competition part of the IEEE MLSP 2025 conference. NOCASA challenges participants to develop new systems that can assess single-word pronunciations of young second language (L2) learners as part of a gamified pronunciation training app. To achieve this, several issues must be addressed, most notably the limited nature of available training data and the highly unbalanced distribution among the pronunciation level categories.
To expedite the development, we provide a pseudo-anonymized training data (TeflonNorL2), containing 10,334 recordings from 44 speakers attempting to pronounce 205 distinct Norwegian words, human-rated on a 1 to 5 scale (number of stars that should be given in the game).   
In addition to the data, two already trained systems are released as official baselines:
an SVM classifier trained on the ComParE\_16 acoustic feature set and a multi-task wav2vec 2.0 model. The latter achieves the best performance on the challenge test set, with an unweighted average recall (UAR) of 36.37\%.

\end{abstract}
\section{Introduction}
\label{sec:intro}

Learning the pronunciation of a foreign or second language (L2) requires a lot of practice and accurate feedback. Mobile apps and online systems that integrate the automatic pronunciation assessment (APA) technology enable learners to practise pronunciation at their own time, place, and pace have gained popularity over the past few years~\cite{10214276,kheir-etal-2023-automatic}. However, these apps and services require an accurate APA model in their backend, which can assess recordings based on the speaker's pronunciation. To build such models, high-quality annotated data and extra care are needed in selecting the right technique and training the models properly. 

Compounding the previously mentioned problems is the fact that children's speech data is considerably rarer than adult corpora. Most available corpora are for the English Language: CMU Kids~\cite{cmukids}, CSLU Kids~\cite{cslukids}, My Science Tutor~\cite{myst};  while some other languages also have some limited resources like the Dutch corpus introduced in~\cite{jasminlr}; the Mandarin data used in~\cite{slt21lr} and the German dataset presented in~\cite{kidstalk}. Despite the progress made in the collection of public corpora, the Norwegian language still lacked any such public data until very recently. 
An additional problem is that most datasets contain only native speakers' recordings and lack L2 speech, which is needed for both the development and testing of APA systems. Fortunately, during the international TEFLON project\footnote{\url{https://teflon.aalto.fi/}}, we were able to collect, annotate and make publicly available a sizeable corpus of children learning the Norwegian language~\cite{TeflonDataLREC2024}. The data contains single-word pronunciations, along with expert ratings, which makes it a prime candidate for training APA solutions and benchmarking.   

In this paper, we introduce a data competition ”Non-native Children’s Automatic
Speech Assessment” (NOCASA), designed to advance automatic pronunciation assessment for young second language (L2) learners of Norwegian, based on the dataset presented in~\cite{TeflonDataLREC2024}. The competition provides a standardized benchmark to evaluate and compare systems for assessing L2 children's pronunciation from single-word speech utterances. 

The NOCASA competition challenges teams to develop automatic speech assessment systems and use them to predict the score of each utterance in the test data. It can be summarized as an unbalanced audio classification problem, although it can also be converted into a regression task. Participants are tasked with developing solutions that can assign a class label (number of stars) to each audio sample uttered by young children. To ensure a comprehensive evaluation, we employ multiple evaluation metrics, including Unweighted Average Recall (UAR), Accuracy, and Mean Absolute Error (MAE). In addition to the official evaluation metrics, we offer some additional considerations, such as model inference speed and explainability, as insights to support further analysis and system development.

\section{Challenges of the Competition}
\label{sec:goal}

Developing and implementing such APA models has several challenges. First and worst is the lack of speech data for L2 learners that would be annotated for pronunciation accuracy. This is particularly the case of children and learners of low-resource target languages. Second, if such data were available, they were usually heavily unbalanced for different skill levels, and the provided reference scores suffer from noise and inter-annotator disagreement. Finally, to make a useful app, the scoring has to happen quickly to provide real-time feedback with minimal delay to encourage re-tries and keep the users engaged.   

\section{Rules of the Challenge}
\label{sec:rules}

Participants must adhere to the predefined training and test splits as given. Additional training data can be used, but only if it is clearly mentioned in the report and the data are publicly available.
The developed solutions could be tested via a maximum of five trials by uploading the model predictions to the test set, the labels of which are unknown to the participants.
Each participation has to be accompanied by an article presenting the results that undergoes a standard double-blind peer-review process, and only authors of accepted papers will be added to the final leaderboard.

Contributors are encouraged to find their own features and use their own machine learning algorithms. However, to enable faster development, a standard feature set for traditional ML models (SVM) and a state-of-the-art end-to-end solution (wav2vec 2.0) is provided (see section~\ref{sec:baselines}).

The organisers reserve the right to re-evaluate the findings but will not participate in the Challenge themselves. We encourage contributions aiming at the highest performance, surpassing our baselines, and contributions aiming at finding new and interesting insights with respect to the shared data.

\section{Data}
\label{sec:data}

\begin{figure*}[t]
    \centering
    \includegraphics[width=0.85\linewidth]{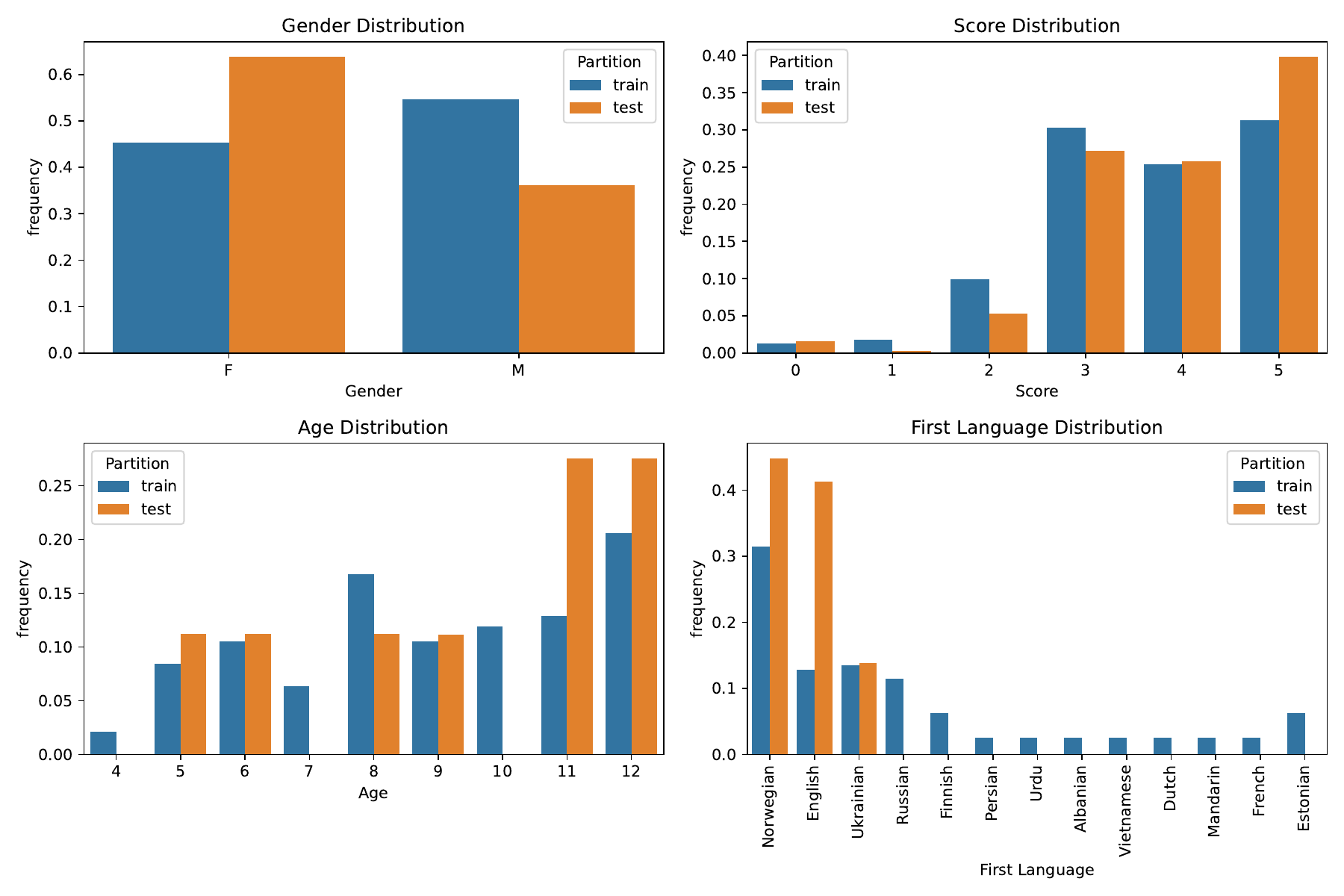}
    \caption{Distribution of relevant speaker characteristics for the training and test set.}
    \label{fig:data_distribution}
\end{figure*}

\begin{table}[]
    \centering
    \resizebox{\columnwidth}{!}{
    \begin{tabular}{cc|cc|cc|cc|cc}
    \hline\hline
    ph & occ & ph & occ &  ph & occ &  ph & occ &  ph & occ \\
    \hline
\textipa{ r } & 83 & 	\textipa{ \:t } & 24 & 	\textipa{ g } & 17 & 	\textipa{ O } & 11 & 	\textipa{ o: } & 4 \\ 
\textipa{ @ } & 79 & 	\textipa{ \:n } & 22 & 	\textipa{ y: } & 15 & 	\textipa{ u: } & 10 & 	\textipa{ E } & 4 \\
\textipa{ l } & 61 & 	\textipa{ \o: } & 20 & 	\textipa{ A: } & 15 & 	\textipa{ i: } & 9 & 	\textipa{ \c{c} } & 3 \\
\textipa{ s } & 51 & 	\textipa{ p } & 20 & 	\textipa{ v } & 15 & 	\textipa{ \oe } & 9 & 	\textipa{ \ae I } & 2 \\ 
\textipa{ k } & 50 & 	\textipa{ f } & 19 & 	\textipa{ m } & 15 & 	\textipa{ h } & 6 & 	\textipa{ \:n } & 1 \\
\textipa{ t } & 43 & 	\textipa{ I } & 18 & 	\textipa{ Y } & 14 & 	\textipa{ \ae } & 6 & 	\textipa{ \:s } & 1 \\
\textipa{ A } & 26 & 	\textipa{ \ae: } & 18 & 	\textipa{ 0 } & 13 & 	\textipa{ N } & 5 & 	\textipa{ E0 } & 1 \\ 
\textipa{ n } & 26 & 	\textipa{ S } & 18 & 	\textipa{ d } & 13 & 	\textipa{ U } & 5 & 	\textipa{ \s{n} } & 1 \\
\textipa{ b } & 25 & 	\textipa{ 0: } & 17 & 	\textipa{ j } & 11 & 	\textipa{ e: } & 4 & 	\textipa{ \oe Y } & 1 \\
    \hline\hline
    \end{tabular}
    }
    \caption{Phonetic inventory with number of occurrences over pronunciations.}
    \label{tab:phoneme_frequency}
\end{table}

The released corpus contains recordings of children of 5 – 12 years repeating one out of 205 words in Norwegian that were played to them.
The children included L1 speakers, beginner learners of L2 Norwegian, and children without previous exposure to Norwegian.
For each word in the training set, we provide orthographic transcription and a speech accuracy assessment score in the range 1--5 given by human experts.
For the test set, we only provide speech and orthographic transcriptions.
The words were chosen to be familiar to Norwegian children, but also to give good coverage of the phonetic inventory.
Table~\ref{tab:phoneme_frequency} shows a list of IPA phonetic symbols that occur in the canonical pronunciation of the 205 unique words, ordered by their frequency of occurrence.
For more general details about data collection, and the assessment criteria, see~\cite{TeflonDataLREC2024}.

The partitioning of the data between training and test set was performed using a controlled randomization procedure.
The goal was to maintain a similar distribution of speech assessment scores, genders, ages, and language backgrounds.
This goal could only be approximated given the relatively low number of speakers.
The initial training set consists of 44 speakers and 10,334 utterances, and the starting test set consists of 8 speakers and 1,930 utterances. To further clear the data, we opted to remove the recordings that have a rating of 0 (i.e. unintelligible, noisy and completely silent samples). Lastly, some children were recorded multiple times saying the same word. To ensure equal representation of each speaker, we chose to include only one recording per word for each speaker. 
Both the training and test sets included the same 205 words, but not all participants uttered all the words.
Overall, in the final released data we shared 7,857 training samples and 1,460 test audio files.
All recordings were between 5 and 8 seconds long and there was no noticeable difference in length between the training and test set recordings.
Figure~\ref{fig:data_distribution} compares the distributions of the relevant speaker characteristics between the training and test set.
It is worth to note that, given the relatively low number of speakers in the test set, only the most common linguistic backgrounds were included, compared to the training set.

As part of the challenge, the file names were randomized to hide the speaker identity.
Although the speaker characteristics were not included in the dataset, we will make this information available when the challenge is completed at the Norwegian Language Bank\footnote{\url{https://www.nb.no/sprakbanken/ressurskatalog/oai-nb-no-sbr-94/}}.

\section{Baselines}
\label{sec:baselines}

We provide an official challenge repository\footnote{\url{https://github.com/aalto-speech/nocasa-baselines}} for reproducing our baseline systems. The repository includes scripts for data preprocessing, model training, and evaluation.

We present two official baseline systems. First, we implement a traditional pipeline approach of a Support Vector Machine (SVM) classifier~\cite{fan-2008-linSVC} trained with the standard ComParE\_16 feature set (6,373 OpenSMILE features)~\cite{schuller16_interspeech}. The task was to choose one of the five possible labels, so we opted for the one-vs-rest classifier strategy during training. To account for the highly unbalanced class distribution, we utilized the class-weighting technique~\cite{king2001logistic} to ensure high performance even for underrepresented classes. As the challenge data did not contain any development sets, we used the default parameters for SVM training, which means that hyperparameter tuning with a cross-validation setup could easily lead to improved systems. This baseline system, by design, is interpretable; thus, the final decisions of the model can be easily explained. Using the SVM weights, we conducted a thorough investigation into which features have been found important for predicting different levels of pronunciation.

Next, we train a multi-task wav2vec 2.0 model (\textit{MT w2v2} in Table~\ref{tab:res_baselines}) following~\cite{getman23_slate}. As the base model, we use a Norwegian ASR wav2vec 2.0 model\footnote{\url{https://huggingface.co/NbAiLab/nb-wav2vec2-300m-bokmaal}}~\cite{de-la-rosa-etal-2023-boosting}. We retain the original ASR (CTC) head and initialize a speech pronunciation rating head on top of the last Transformer layer, consisting of a linear projection, average pooling, and a classification layer. We update all model weights except for the convolutional feature encoder, which remains frozen.

The model is trained in a multi-task setting for 20 epochs with a batch size of 10 and a peak learning rate of 2e-4. For hyperparameter tuning, we hold out 10\% of the data as a validation set, then re-train the final model on the full dataset and select the checkpoint corresponding to the best epoch from the previous training run. Since only the expected (target) word is known for each speech sample, rather than the actual pronounced word, we include target words as ASR references during training only when the pronunciation score is 4 or 5.

Although the last layer of wav2vec 2.0 may not be the most optimal for the speech rating task, we adopt it as our baseline without performing a layer-wise analysis. Additionally, in this baseline, we do not leverage ASR head predictions to verify whether the pronounced word corresponds to the target word or to adjust the speech rating head's predictions.

Table~\ref{tab:res_baselines} presents the results of the baseline systems on the challenge test set. As the primary evaluation metric, we use unweighted average recall (UAR). Additionally, we measure accuracy (ACC) and mean absolute error (MAE).
For UAR, we also report 95\% confidence intervals (CIs) computed via bootstrapping~\cite{CI}. We achieve the best UAR of 36.37\% on the test set with the multi-task wav2vec 2.0 model. We also evaluate the ASR performance of the multi-task system on utterances with a pronunciation score of 4 or 5, obtaining a WER of 10.63\% and a CER of 4.10\%.

\begin{table}
    \centering
      \begin{tabular}{lccc}
         \toprule
       \textbf{System} & \textbf{UAR, \%} & \textbf{ACC, \%} & \textbf{MAE} \\
       \hline
         SVM & 22.14 (20.45 -- 24.45) & 32.74 & 1.05 \\
         MT w2v2 & 36.37 (34.30 -- 38.81) & 54.45 & 0.55 \\
         \hline
         \hline
         Team 1 & 42.52 (33.13 -- 53.80) & 45.14 & 0.68 \\
         Team 2 & \textbf{44.81 (38.99 -- 53.74)} & \textbf{55.75} & \textbf{0.50}\\
         \bottomrule
      \end{tabular}
    \caption{Unweighted average recall (with 95\% CIs), accuracy and mean absolute error of baseline models on the test set, and results of the top two participants.}
    \label{tab:res_baselines}
\end{table}

\section{Discussion}
\label{sec:discuss}

While high performance is necessary for the usability of SLA systems, it is not the only measure that should be considered. For usability, one should also ensure that the overall system's latency is not too high, otherwise the users, especially young children, would abandon the game. To adhere to this criterion, we opted to use the \textit{large} size model, with 300M parameters, and measure its inference speed. On average, the baseline wav2vec2 model proved to be the slowest, requiring approximately 30-50 milliseconds to process a 3-second-long audio file using a single GPU, meaning that its overall latency is within an acceptable range (less than a second).

A second aspect to consider is model robustness and transparency. To test the robustness of our best wav2vec 2.0 model, we trained 2 more variants with different random seeds to see how stable its performance is. In terms of accuracy, the models varied between 52.9\% -- 55.8\%, while their UAR values were between 36 and 38, well within the CI interval of Table~\ref{tab:res_baselines}. Looking at the MAE values (between 0.53 -- 0.57), we concluded that the 3 models made only slightly different minor mistakes, leading to an overall conclusion that our system is robust to random seed changes. 

We should note that the choice to use multiple evaluation metrics was motivated by the fact that they offer a more nuanced understanding of the models' performance. The standard accuracy metric is suitable to measure how successfully the models learned the discrimination task in general, but it could be misleading, as systems overfitting to the majority classes could offer high accuracy, but would mostly predict 3+ stars. In contrast, MAE could offer us an insight into the difference between the predicted and actual labels, thus penalizing large mistakes, thus complementing the accuracy metric. Lastly, the UAR considers the models' performance (Recall) per category, which offers potential users information on how well the model, in general, would work at any level. UAR is also less sensitive to data imbalance, unlike the other two metrics, due to the unweighted averaging of Recalls, giving equal importance for good performance on rare categories and majority classes. Hence, we opted to select it as the main metric.

Next, we investigated our systems' functionality. While our primary user group (young language learners) would not benefit from being exposed to detailed reasons behind the automated system's decisions, we argue that secondary stakeholders (parents and teachers) might benefit from these insights. First we investigated our SVM models, which by their nature are easy to explain. To ensure that the investigation of the coefficients paints the true picture, we decided to use a feature normalisation step in the pipeline; thus, the coefficients of each input feature are directly comparable. Our investigation revealed that the model considered the distances between MFCC peaks and the features related to rising and falling MFCC slopes to be important for predicting low categories ($\leq$ 3 stars), while the model predicting the highest categories valued the PCM RMS energy statistics and some LPC features extracted by audSpec. These observations, combined with some knowledge in phonetics, could lead to valuable feedback on how users should try to change their pronunciation if they wish to achieve higher scores.

\begin{figure}[t!]
    \centering
    \includegraphics[width=0.9\linewidth]{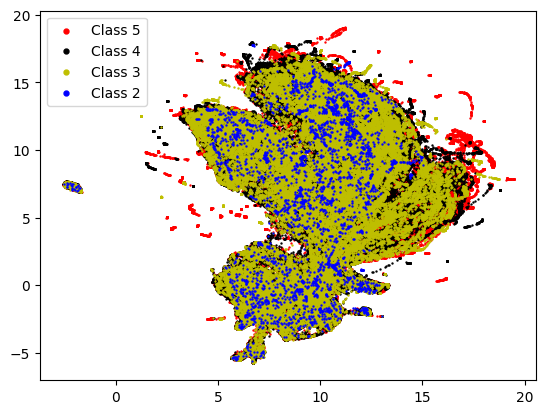} \includegraphics[width=0.9\linewidth]{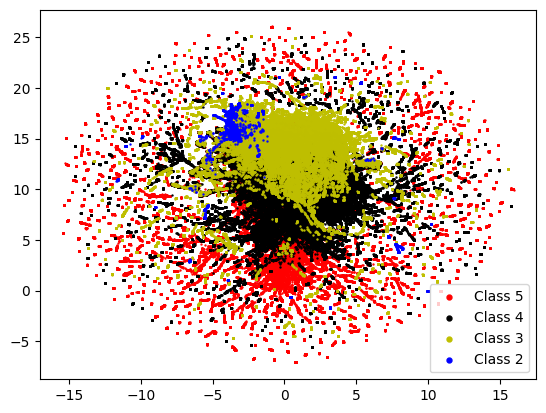}
    \caption{UMAP visualization of the last Transformer layer's outputs on the test set before (top) and after (bottom) fine-tuning, colors assigned based on the predicted labels.}
    \label{fig:vis}
\end{figure}

Regarding the wav2vec 2.0 model, we conducted a thorough investigation including latent embedding visualisation using the UMAP~\cite{healy2024uniform} algorithm. During this analysis, we refrained from using the gold-standard test labels, and instead used the labels predicted by our model (\textit{MT w2v2}), to demonstrate what kind of analysis participants could do. First, we compared the models' embeddings before and after the fine-tuning procedure to see if we could observe any of the expected groups forming. Figure~\ref{fig:vis} allows us to visually inspect the embeddings produced by the last Transformer layer, which was directly used by the classification head of the model. We can observe that before fine-tuning, the model does not differentiate between various pronunciation levels, and after the fine-tuning procedure, the clusters start to form, but they are not yet well separated. We hypothesise that mainly the lack of training data and the limited number of training steps lead to this non-optimal embeddings space, which is reflected in the low performance of the model. We would like to point out that this analysis also revealed a model weakness, namely that the lowest level (1 star), which is the rarest in the training data, is never predicted. We strongly encourage participants to address this issue by using training techniques designed for unbalanced data. 

Next, we turned our attention to the model confidence, as end-to-end speech classifiers are known to be over-confident \cite{hao24_interspeech}, which leads to problems during inference. On the other hand, low probability predictions are usually a sign of sub-optimal training and signal that additional hyperparameter tuning is required. On the test data, we observed that the average probability of the predicted category was approximately 84\%, which signals that the model in general was quite confident about its decisions, but the high standard deviation ($\pm$16\%) indicates that it was quite often conflicted by some samples. Based on these observations, we can conclude that the model has been trained well, but would benefit from additional post-processing, such as posterior calibration or converting to a hierarchical classification approach.

\section{Submitted Solutions}
\label{sec:solutions}

In this section, we provide a brief overview of the solutions submitted by NOCASA participants that outperformed our baselines. 

Team 1 has opted to solve the challenge via multitask regression models~\cite{team1}. By changing the training objective, they allowed the models to differentiate between proficiency levels within the same scoring category. For example, category 3 could mean almost 4 or slightly better than 2, which are quite different. Their solution with an additional score calibration step managed to outperform both baselines (see Table~\ref{tab:res_baselines}) and has the potential to detect small improvements in pronunciation proficiency.

Team 2 approached the problem from a different angle; they compared several end-to-end models and explored ways of converting their CTC emission probabilities to goodness-of-pronunciation (GOP) scores~\cite{team2}. In addition, they integrated information about the word that the child is attempting to pronounce via prefix-tuning. Lastly, to address the ordinal nature of the labels, they utilized a weighted ordinal Cross-Entropy loss, which proved to be the winning solution, achieving first place on the leaderboard (see Table~\ref{tab:res_baselines}).

\section{Conclusions}

In this paper, we introduced the "Non-native Children's Automatic Speech Assessment" (NOCASA) competition, aiming to offer a common benchmark for APA solutions. Along with the training data, we have released two baseline solutions: a traditional pipeline of acoustic feature extractions combined with an SVM model and an end-to-end wav2vec 2.0 system. The best official baseline system achieved 36.37\% UAR on the test data, leaving plenty of room for improvement. We expect participants to obtain better results by changing the model architecture or by using various training strategies designed for unbalanced data, optimising their hyperparameters, and exploiting the extra information (expected word to be pronounced) in different ways compared to our baseline. We strongly encourage our participants to consider and balance their solutions' latency and performance across the official evaluation metrics.

\section{Acknowledgements}
\label{sec:acknowledgements}

This work was supported by NordForsk through funding to technology-enhanced foreign and second-language learning of Nordic
languages (TEFLON) under Project 103893. The computational resources were provided by Aalto
ScienceIT.

\printbibliography

\end{document}